\documentclass[conference]{IEEEtran}
\IEEEoverridecommandlockouts
\usepackage{cite}
\usepackage{amsmath,amssymb,amsfonts}
\usepackage{algorithmic}
\usepackage{mathtools}
\usepackage{graphicx}
\usepackage{textcomp}
\usepackage{xcolor}
\usepackage{booktabs}
\usepackage{multirow}
\usepackage{enumitem}  
\usepackage{verbatim}
\usepackage{subfigure}
\usepackage[font=small,skip=10pt]{caption}
\usepackage{hyperref}
\usepackage{booktabs}
\newcommand{\tabitem}{~~\llap{\textbullet}~~}

\def\BibTeX{{\rm B\kern-.05em{\sc i\kern-.025em b}\kern-.08em
    T\kern-.1667em\lower.7ex\hbox{E}\kern-.125emX}}
\begin{document}
\newcommand{\zc}[1]{\textcolor{red}{\textbf{ZC:} #1}}
\newcommand{\jiamou}[1]{\textcolor{blue}{#1}}

\definecolor{aqua}{rgb}{0.0, 1.0, 1.0}
\definecolor{mygreen}{RGB}{0, 176, 80}
\definecolor{mylightorange}{RGB}{255, 192, 0}
\definecolor{mypurple}{RGB}{112, 48, 160}
\definecolor{myblue}{RGB}{0, 112, 192}
\definecolor{mypink}{RGB}{255, 102, 255}
\definecolor{mybrown}{RGB}{127, 96, 0}

\title{Generating Informative CVE Description From ExploitDB Posts by Extractive Summarization\vspace{-3mm}}

\author{\vspace {-5mm}
	\IEEEauthorblockN{Jiamou Sun\IEEEauthorrefmark{1},  Zhenchang Xing\IEEEauthorrefmark{1}, Hao Guo\IEEEauthorrefmark{2}, Deheng Ye\IEEEauthorrefmark{3},  Xiaohong Li\IEEEauthorrefmark{2},Xiwei Xu\IEEEauthorrefmark{4}, Liming Zhu\IEEEauthorrefmark{4}}
	\\
	\IEEEauthorblockA{\IEEEauthorrefmark{1}Research School of Computer Science, CECS, \textit{Australian National University}, Canberra, Australia} 
	\IEEEauthorblockA{\IEEEauthorrefmark{2}College of Intelligence and Computing, \textit{Tianjin university}, Tianjin, China } 
	\IEEEauthorblockA{\IEEEauthorrefmark{3}Tencent AI Lab, Shenzhen, China} 
	\IEEEauthorblockA{\IEEEauthorrefmark{4}Data61, CSIRO, Australia}
	\vspace {-4mm}
	\\ \{u5871153, Zhenchang.Xing\}@anu.edu.au, \{haoguo, xiaohongli\}@tju.edu.cn 
	\\ dericye@tencent.com, \{Xiwei.Xu, Liming.Zhu\}@data61.csiro.au
	\vspace {-7mm}}

\maketitle

\begin{abstract}
	
	ExploitDB is one of the important public websites, which contributes a large number of vulnerabilities to official CVE database.
	Over 60\% of these vulnerabilities have high- or critical-security risks.
	Unfortunately, over 73\% of exploits appear publicly earlier than the corresponding CVEs, and about 40\% of exploits do not even have CVEs.
	To assist in documenting CVEs for the ExploitDB posts, we propose an open information method to extract 9 key vulnerability aspects (vulnerable product/version/component, vulnerability type, vendor, attacker type, root cause, attack vector and impact) from the verbose and noisy ExploitDB posts.
	The extracted aspects from an ExploitDB post are then composed into a CVE description according to the suggested CVE description templates, which is must-provided information for requesting new CVEs.
	Through the evaluation on 13,017 manually labeled sentences and the statistically sampling of 3,456 extracted aspects, we confirm the high accuracy of our extraction method.
	Compared with 27,230 reference CVE descriptions. Our composed CVE descriptions achieve high ROUGH-L (0.38), a longest common subsequence based metric for evaluating text summarization methods.
	
\end{abstract}
\section{Introduction}
\label{sec:intro}

Vulnerabilities in computer software and hardware are inevitable.
Once discovered, vulnerabilities should be systematically documented and managed, which is crucial for the practitioners and researchers to develop patches and prevention mechanisms~\cite{Chaowei@2018, Wijayasekara@ieee2012, Sabottke@usenix2015, Feng@usenix2019}.
Common Vulnerabilities and Exposures (CVE) is the most influential vulnerability database, managed by the non-profit organization Mitre~\cite{CVE}.
As of August 2020, CVE database documents 185,660 vulnerability entries.
CVE entries are used in numerous security products and services, including the US National Vulnerability Database (NVD)~\cite{NVD}.

\begin{figure}
	\centering
	\includegraphics[width=0.95\linewidth]{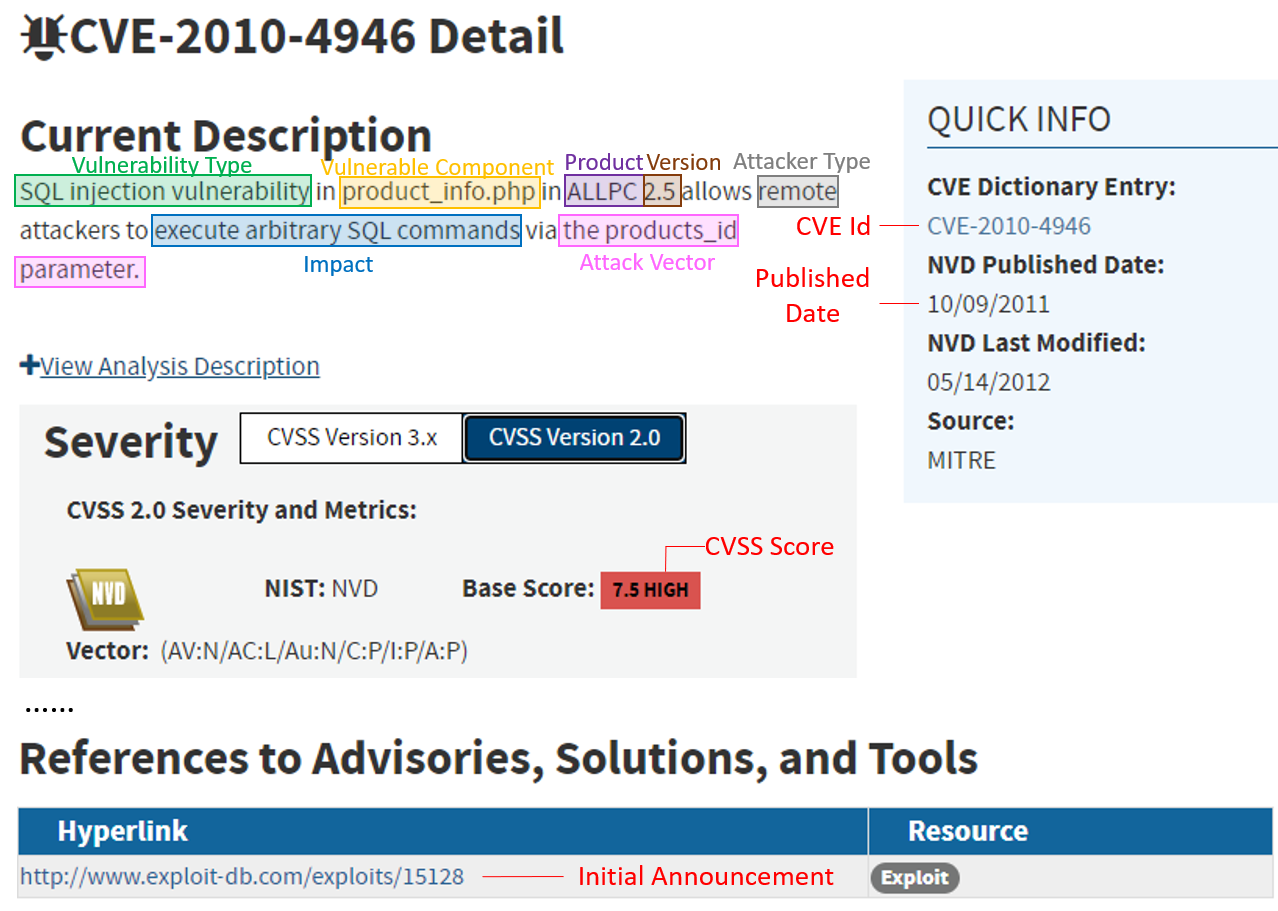}
	\vspace{-3mm}
	\caption{A CVE Entry \href{https://nvd.nist.gov/vuln/detail/CVE-2010-4946}{CVE-2010-4946} in the NVD Database}
	\label{fig:exp_nvd}
	\vspace{-6mm}
\end{figure}

Fig. \ref{fig:exp_nvd} shows a CVE entry in the NVD.
Each CVE entry has an identification number, for example, \href{https://nvd.nist.gov/vuln/detail/CVE-2010-4946}{CVE-2010-4946} for the CVE entry in Fig.~\ref{fig:exp_nvd}.
Vulnerability description is a must-provided information.
It should describe various aspects of the vulnerability, such as vulnerability type, vulnerable component in which product and version(s), attacker type, impact, and attack vector.
We highlight these aspects in the description of the CVE-2010-4946 in Fig.~\ref{fig:exp_nvd}.
NVD analyses the CVE entries and assigns Common Vulnerability Severity Scores (CVSS)\footnote{CVSS score is calculated based on various exploitability and impact metrics. It ranges from 0 to 10, with four severity levels - Critical (9.0-10), High (7.0-8.9), Medium (4.0-6.9), Low (0-3.9). In this work, we use CVSS 3.x score if it is available, other CVSS 2.0 score.}.
CVE-2010-4946 is assigned a HIGH-severity score.
According to the CVE documentation convention~\cite{CVEConven}, the first reference is the initial announcement of the vulnerability.
In this example, the initial announcement is an exploit entry submitted to the \href{https://www.exploit-db.com/}{ExploitDB website}~\cite{ExploitDB} on 27 September 2010.

ExploitDB is one of the most referenced public websites for official CVEs (see Section~\ref{sec:exploitimportance}).
69\% of CVEs whose initial announcements come from the ExploitDB are of HIGH- or CRITICAL-severity.
Unlike many vendor-and-project sources (e.g., IBM X-Force), ExploitDB is contributed by ``the crowd''.
Anyone who discovers some new exploits of any products can publish the exploits on the ExploitDB website.
However, they have no obligations to submit the discovered exploits to product vendors or Mitre (so-called CVE Numbering Authorities, CNA) who are authorized to assign CVE ID to vulnerabilities.
In fact, to request a new CVE ID, the exploit discover has to compile the vulnerability information in the required format~\cite{CVERequest}, and follow a rather sophisticated process to communicate with product vendors and Mitre~\cite{CVE}.

\begin{figure}[tb]
	\centering
	\includegraphics[width=\linewidth]{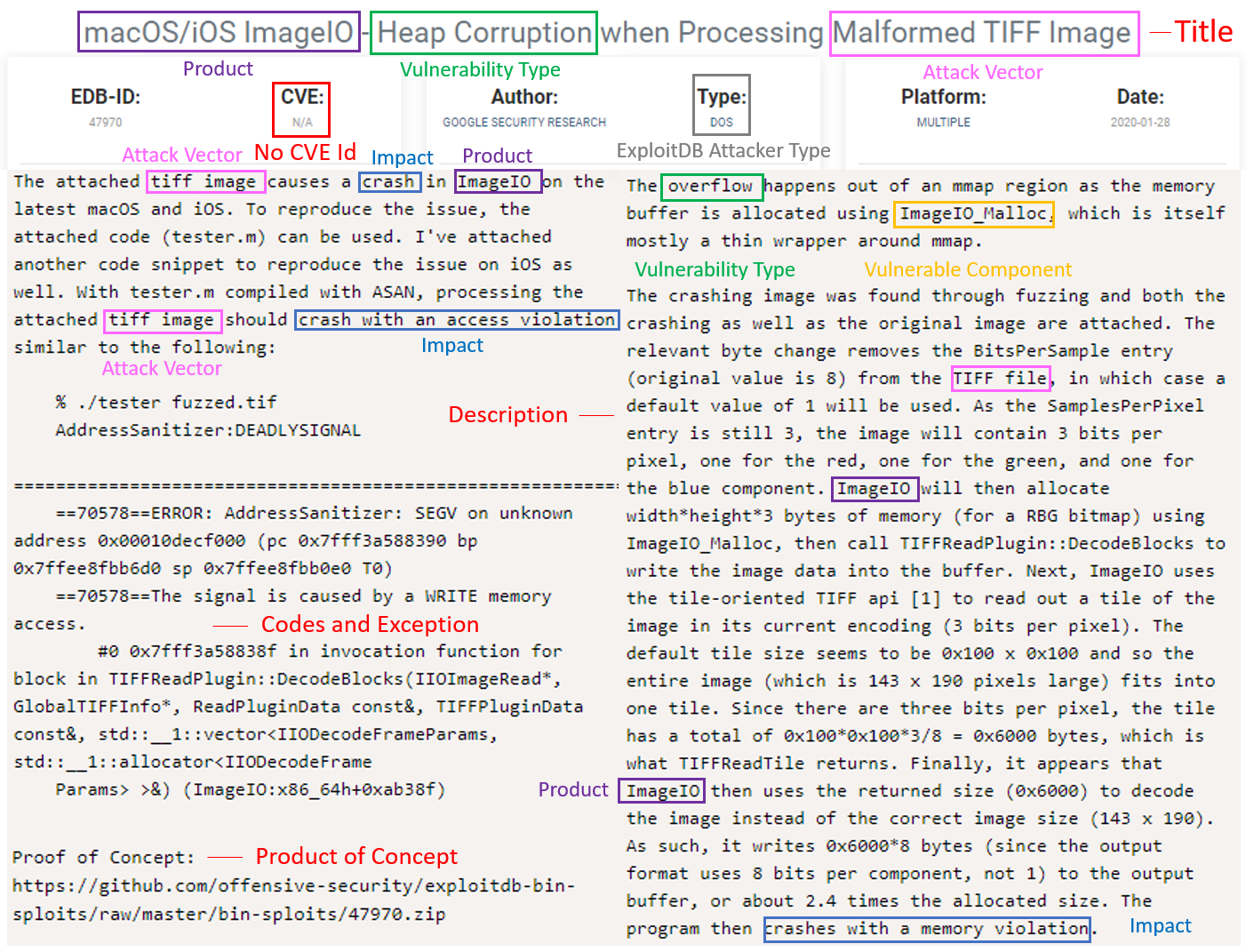}
	\vspace{-6mm}
	\caption{An \href{https://www.exploit-db.com/exploits/47970}{Exploit Post 47970} on the ExploitDB website}
	\label{fig:exp_exploitdb}
	\vspace{-8mm}
\end{figure}

It may take a long time to document a crowd-contributed exploit as an official CVE entry (see Section~\ref{sec:delay}).
For example, CVE-2010-4946, albeit HIGH-severity, was published in the CVE/NVD almost one year after the initial announcement of the exploit on the ExploitDB.
We investigate 41,883 exploit posts (as of December 31 2019) on the ExploitDB.
We find that 73.5\% exploits were announced at least one day earlier than the published date of the corresponding CVEs.
57\% of these CVEs are of HIGH- or CRITICAL-severity.
Even worse, about 28\% of exploits on the ExploitDB do not have corresponding CVE entries (see Section~\ref{sec:lackcve}).
Fig.~\ref{fig:exp_exploitdb} shows such an exploit post \href{https://www.exploit-db.com/exploits/47970}{ExploitDB:47970}. 
This exploit was posted on 28 January, 2020.
It is about the \textit{ImageIO} framework  of the latest \textit{iOS} and \textit{macOS} operating systems (product).
The vulnerable component \textit{ImageIO\_Malloc} suffers from \textit{heap corruption} (vulnerability type) that causes \textit{system crash} (impact) via \textit{malformed TIFF image} (attack vector).
Unfortunately, no CVE entry officially documents this vulnerability even after 7 months of the exploit's publication.
The exploits published on the ExploitDB come with Proof-of-Concepts (PoC) showing how the vulnerabilities can be exploited.
These PoCs aim to help people understand the vulnerabilities and develop patches, but they can also be exploited by adversaries.
The delay and missing in documenting CVEs for these exploits would weaken mitigation and prevention efforts.

This work makes the first effort to address this delay and missing issue.
In particular, we develop an extractive summarization approach that automatically composes a CVE description in the required format~\cite{CVERequest} from an exploit post on the ExploitDB.
As shown in Fig.~\ref{fig:exp_exploitdb}, an exploit is composed of a concise title, an exploit description, a PoC code, and some metadata (e.g., ExploitDB attacker type).
Many aspects of an informative CVE description can be extracted from different parts of the exploit.
However, these aspects may be scattered in different parts, and most of the information is in free-form text and mixed with code.
Furthermore, the same aspect may be mentioned multiple times and may be described in various ways, for example, product name, vulnerability type and attack vector in the example in Fig.~\ref{fig:exp_exploitdb}.
The content structure and description style vary from one exploit to another (compare the example in Fig.~\ref{fig:exp_exploitdb} with a directory-traversal exploit \href{https://www.exploit-db.com/exploits/48232}{VMWare Fusion - Local Privilege Escalation}). 

The complexity of the exploit text renders rule-based extraction method ineffective, except for simple metadata like attacker type and vendor.
Therefore, we adopt machine learning based open information extraction methods to extract candidate CVE aspects from the exploit text.
Considering the lexical gap between the exploit text and the CVE descriptions (see Table~\ref{tab:composedreferenceexamples}), we do not use end-to-end text summarization methods~\cite{Sutskever@2014,Moratanch@2016, Nenkova@2011}.
Instead, we explicitly extract different aspects first and then compose them according to the CVE description templates~\cite{CVEtemplate}.
Based on the data characteristics of different aspects, we use BERT-based Named Entity Recognition (NER)~\cite{BERT} to extract vulnerability product/version/component and vulnerability type, use rule-based NER to extract vendor and attacker type, and use BERT Question Answering (QA) to extract root-cause, attack-vector and impact clauses.

We label the CVE aspects in the 13,017 sentences of 1,764 ExploitDB posts for training and testing the NER and QA models.
BERT-based NER achieves 0.76 in F1, and BERT QA achieves 0.77 exact-matching score and 0.82 partial matching score.
They significantly outperform rule-based extraction methods.
We use statistical sampling method~\cite{singh2013elements} to examine the extracted aspects from over 44K exploit posts.
The accuracy for all aspects but vendor is $>$0.9, and the accuracy for vendor is 0.8.
We use ROUGE-1/2/L metrics \cite{Rouge} to evaluate the similarity of the composed CVE descriptions and reference CVE descriptions.
Our approach outperforms the state-of-the-art text summarization methods~\cite{liu@emnlp2019} on all ROUGE metrics.

The main contributions of this paper are as follows:

\begin{itemize}
	\item We conduct an empirical study of 41,883 ExploitDB posts to understand the importance of the ExploitDB posts for CVE discovery and analysis, and the delay and missing in documenting exploits as CVEs.
	
	\item We propose a NER- and QA-based information extraction approach to extract CVE aspects from an exploit post and compose an informative CVE description for the exploit.
	
	\item We conduct experiments to confirm the accuracy of our information extraction methods and the quality of the composed CVE descriptions.
\end{itemize}

\section{Formative Study}
\label{sec:for_study}

We collect 41,883 exploit posts (as of December 31 2019) on the ExploitDB website.
We conduct an empirical study of these exploit posts to investigate the relationships between the ExploitDB exploits and the CVEs from three aspects:

\begin{itemize}
	\item How important are the ExploitDB exploits for the discovery and analysis of CVEs?
	
	\item Are there significant delays in documenting the ExploitDB exploits as CVEs? What impact could the delay incur?
	
	\item How severe is it that the ExploitDB exploits do not have CVE entries?
	
\end{itemize}


\subsection{The Importance of Exploits for CVEs}
\label{sec:exploitimportance}

To understand the importance of the ExploitDB exploits for the CVEs, we examine the exploits that have been referenced by the CVEs as initial announcements.
This work focuses on the exploits published on the public platforms, because these exploits are open to everyone and much less strictly controlled than those managed by vendor and project CNAs.
Therefore, we exclude the initial announcement websites that come from \href{https://cve.mitre.org/cve/cna.html}{vendor and project CNAs} who are responsible for the vulnerabilities found in their own products and projects.

We collect in total 2,502 initial-announcement websites (excluding vendor/project-CNA websites).
Among these 2,502 websites, the ExploitDB is the third most referenced website, following SecurityTracker~\cite{SecurityTracker} and VUPEN~\cite{VUPEN}.
However,  the SecurityTracker data is only updated to 2018, and the VUPEN website is no longer publicly available.
That is, the ExploitDB is the top-1 referenced public website that still actively contributes initial vulnerability announcements to the CVE.
Among 1,078 CVEs with the ExploitDB exploits as initial announcements, 69\% are of HIGH- or CRITICAL severity (i.e., CVSS scores$\geq$7.0), 30.6\% are of MEDIUM severity, and less than 0.4\% are of LOW severity.

We also collect in total 8,173 non-vendor/project-CNA websites referenced by the CVEs, including initial announcements and all other referenced resources.
ExploitDB is the fifth most referenced website, following SecurityFocus~\cite{Securityfocus}, SecurityTracker~\cite{SecurityTracker}, OSVDB~\cite{OVSDB} and VUPEN~\cite{VUPEN}.
Among 11,642 CVEs that reference the ExploitDB posts, 65\% are of HIGH- or CRITICAL severity, 29\% are of MEDIUM severity, and only about 6\% are of LOW severity.

\subsection{The Delay in Creating CVE Entries} 
\label{sec:delay}

\begin{figure}[bt]
	\centering  
	\subfigure[\% of Exploits Earlier than CVEs by Years]{
		\label{fig:earlier_exploitdb}
		\includegraphics[width=0.23\textwidth]{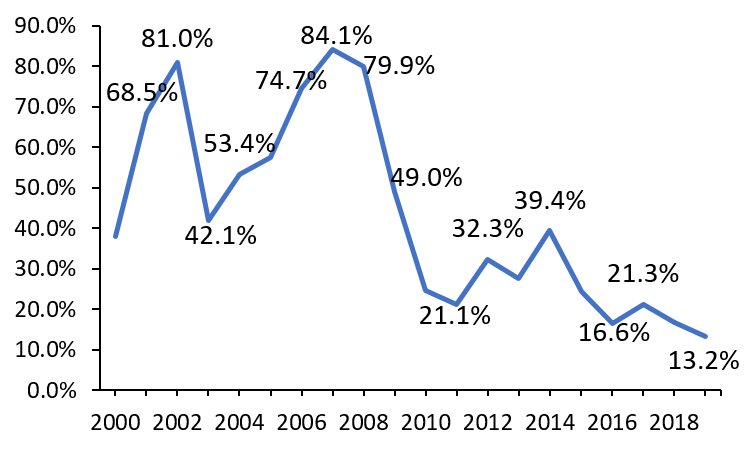}}
	\subfigure[\% of Exploits Earlier (or Later) than CVEs by Day Ranges]{
		\label{fig:exploitdb_earlier_rate}
		\includegraphics[width=0.23\textwidth]{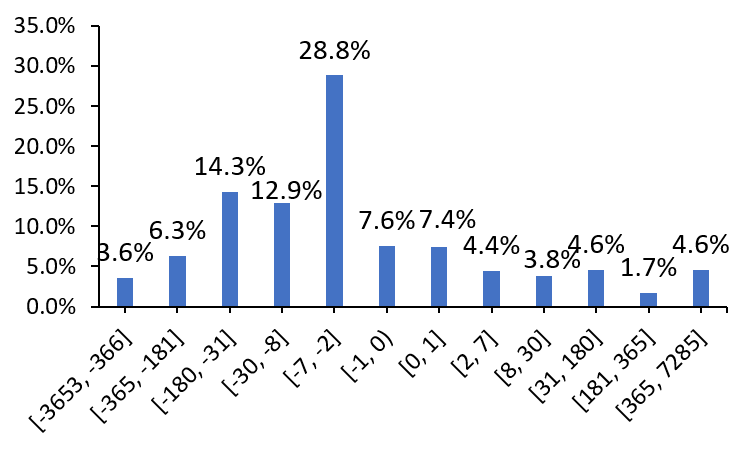}}
	\vspace{-4mm}
	\caption{Statistics of Exploits Earlier than CVEs}
	\label{fig:cve_expl}
	\vspace{-6mm}
\end{figure}

To understand the delay in creating CVE entries for the ExploitDB exploits, we collect the ExploitDB exploits that have CVE IDs.
We compare the published date of the exploits with the published date of the corresponding CVEs to determine if the exploits appear earlier (or later) than the corresponding CVEs.
Fig. \ref{fig:cve_expl}(a) shows the percentage of the exploits published earlier than the corresponding CVEs by years.
We can see a decreasing trend of this percentage over time, which is a good signal.
For example, only 210 (13\%) exploits published in 2019 appear earlier than the corresponding CVEs.
However, 210 is still not a small number, although the percentage is not very high.

Fig.~\ref{fig:exploitdb_earlier_rate} plots the percentage of exploits published earlier (or later) than the corresponding CVEs by six-day ranges [0, 1], [2, 7], [8, 30], [31, 180], [180, 365], and $>$365.
We use negative numbers to represent earlier-than.
For all exploits with corresponding CVEs, the earlier-than exploits account for 73.5\%, which is about 3 times more than the later-than exploits.
For the exploits published earlier than the CVEs, 
57\% of the corresponding CVEs are of HIGH- or CRITICAL severity.
These vulnerabilities could pose significant threats to the vulnerable systems, because the ways to exploit them have been disclosed before the defence becomes available.

For the exploits published earlier than the CVEs, 7.6\% are earlier than 1 day, 28.8\% are within 2 to 7 days.
We see a high percentage 37.1\% of exploits published more than one week earlier than the corresponding CVEs.
There are even 3.6\% of exploits published more than one year earlier.
In an extreme case, a CVE was created in 2020 for the \href{https://www.exploit-db.com/exploits/19943}{ExploitDB:19943} which was published 10 years ago.
The large time gap between the exploits and the CVEs indicates that the risks of some vulnerabilities could be very long-lasting.

\subsection{The Severity of Missing CVE Entries} 
\label{sec:lackcve}

\begin{figure}[bt]
	\centering  
	\subfigure[\% of Exploit without CVE by Years]{
		\label{fig:lackcve_exploitdb}
		\includegraphics[width=0.23\textwidth]{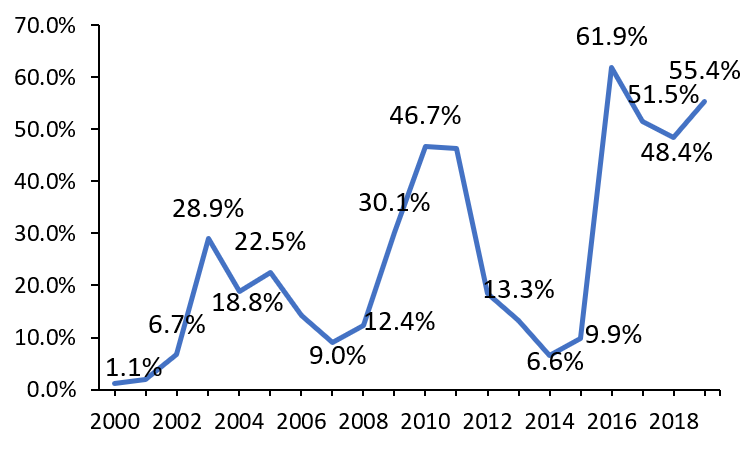}}
	\subfigure[\% of Exploits without CVE by Day Ranges]{
		\label{fig:exploitwithoutcvebydayranges}
		\includegraphics[width=0.23\textwidth]{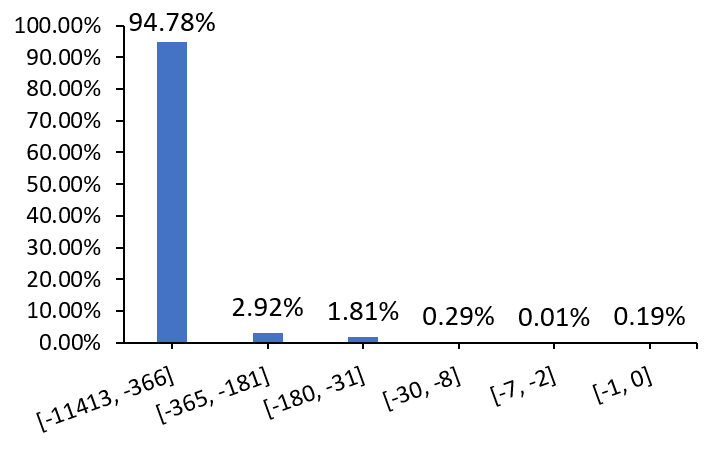}}
	\vspace{-4mm}
	\caption{Statistics of Exploits without CVE}
	\label{fig:cve_exp2}
	\vspace{-6mm}
\end{figure}

We determine if an exploit has the CVE by checking the exploit's CVE metadata.
Among the 41,883 exploits, 16,604 (39.6\%) exploits do not have corresponding CVEs (i.e., their CVE metadata is null, see Fig.~\ref{fig:exp_exploitdb} for an example).
We calculate the number of days elapsed since the published date of these 16,604 exploits till December 31 2019.
Fig.~\ref{fig:exploitwithoutcvebydayranges} shows the distribution of the number of days elapsed.
We still use negative numbers to represent earlier-than.
94.8\% exploits are more than 1 year old since their published date, and 4.7\% exploits are 1 month to 1 year old.
It seems that if an exploit were not documented timely, the chance that it will be documented in the future diminishes as time passes.

Fig.~\ref{fig:lackcve_exploitdb} shows the percentage of the exploits without CVE by years.
We observe two fluctuations in the past 20 years.
The increasing of the exploits without CVEs over time seems to raise the community's attention to reduce such exploits, for example, from 2004 to 2007 and from 2012 to 2014.
But the overall trend of the exploits without CVEs has been rising up, and we are now in a rising period (2016-2019).
For example, 55.4\% of the exploits published in 2019 do not yet have CVE. 
For the exploits without CVEs, we cannot judge their severity.
However, considering that 57\% of the exploits with CVEs are of HIGH- or CRITICAL-severity, these exploits without CVEs could pose significant security threats.

\vspace{1mm}
\noindent\fbox{\begin{minipage}{8.6cm} \emph{The ExploitDB exploits is the top-1 active public resources for CVE discovery and analysis. Unfortunately, there is a significant delay in documenting the ExploitDB exploits as CVEs, the majority of which have HIGH- or CRITICAL severity. Even worse, there is a large portion of exploits without CVEs at all. Although the trend of delay has been going down, the trend of the exploits without CVEs has been rising up in the recent years. Considering the overall severity of the ExploitDB exploits as well as their potential long-lasting risks, it is worth reducing the exploits without CVEs and the delay in documenting them.} \end{minipage}}\\

\section{The Approach}
\label{sec:method}

To mitigate the delay and missing in documenting CVEs for the ExploitDB posts, the first step is to facilitate the generation of CVE descriptions from detailed exploit posts.
We support this step by an extractive text summarization approach.
Different from existing black-box text summarization methods~\cite{Rouge}, our approach adopts a white-box strategy.
Based on the characteristics of different CVE aspects, we adopt named entity recognition and question-answering techniques to extract 9 key aspects from the exploit posts (Section~\ref{sec:extraction}), and then compose the extracted aspects according to the CVE templates~\cite{CVERequest} recommended by Mitre (Section~\ref{sec:compose}).

\begin{figure*}
	\centering
	\includegraphics[width=0.95\linewidth]{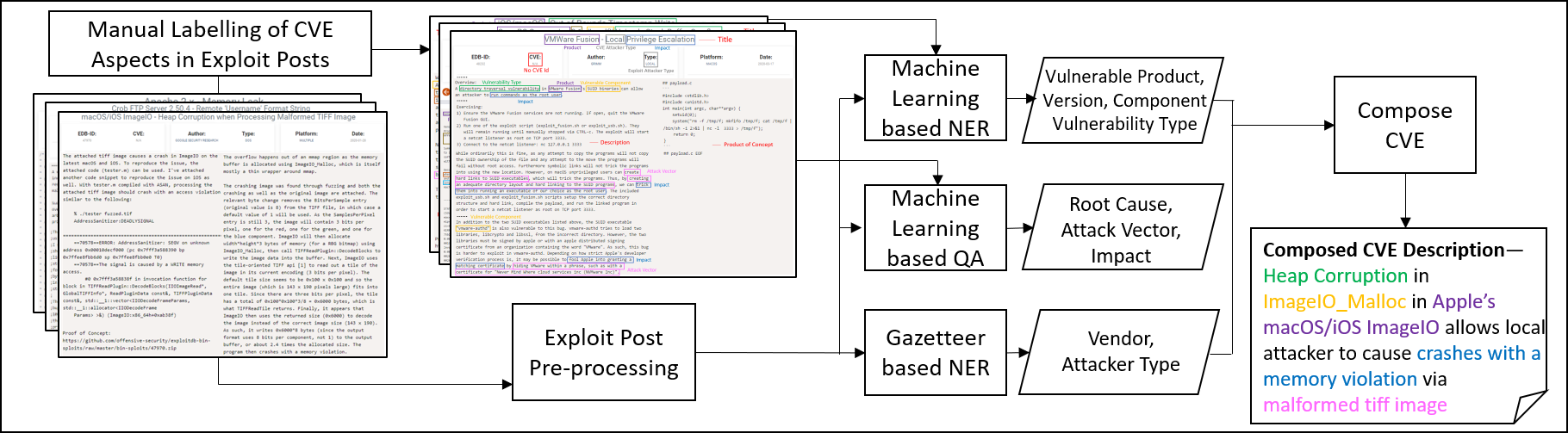}
	\vspace{-2mm}
	\caption{Approach Overview}
	\label{fig:method_overview}
	\vspace{-6mm}
\end{figure*}

\subsection{Exploit Post Pre-processing}
\label{sec:preprocessing}
Given an ExploitDB post, we remove the proof-of-concept information which mainly includes code snippets and exception information.
We use the Spacy tool~\cite{spacy} to split the description text into sentences, and then use the software-specific tokenizer~\cite{dehengicsme2016, apcaveatkgicsme2018} to tokenize the sentences.
This tokenizer has been developed for parsing API documents and Stack Overflow posts, which is similar to the ExploitDB posts.

\subsection{Extracting CVE Aspects from Exploits}
\label{sec:extraction}

According to the CVE description templates~\cite{CVEtemplate} recommended by Mitre, an informative CVE description consists of 9 key aspects, including vendor, vulnerable product, version(s) and component(s), vulnerability type or root cause, impact, attacker type, and attack vector.
We extract these aspects from the exploit post using machine-learning based named entity recognition (NER), gazetteer (i.e., dictionary)-based rules, and question-answering based clause extraction.

\subsubsection{Extracting Vulnerable Product, Version, Component and Vulnerability Type by BERT-based NER}

Each vulnerability usually affects a single product, but may affect multiple components and versions.
Products are identified by their names, such as \emph{ImageIO}, \emph{iOS} and \emph{macOS} in Fig.~\ref{fig:exp_exploitdb}.
As the exploits are described in free-form text, the same product may be mentioned in various forms, including abbreviations, synonyms and even misspellings, such as \textit{Microsoft Windows 10} or \textit{Win10}. 
Another issue is that product names may also be common words.
For example, \emph{chat} can mean a chatting software (e.g., \emph{X7 chat}), or it can mean talking to someone.
This is referred to as common word polysemy~\cite{dehengicsme2016}.

Versions are number sequences like \textit{2.5.2}.
A vulnerability may affect a set of product versions, often described by temporal words before, after or prior, such as \textit{before 2.5.2}.
These temporal words are also polysemous.
They can be used to describe non-version meanings.
For example, in the sentence ``after all, it is an MS product'', the word \emph{after} is not about the temporal order, and it is not followed by versions.
Vulnerable components are program elements at different granularities, such as executables (e.g., \textit{vmware-authd} in \href{https://www.exploit-db.com/exploits/48232}{ExploitDB:48232}), source files (e.g., \textit{main.swf} in \href{https://www.exploit-db.com/exploits/40106}{ExploitDB:40106}), modules (e.g., \textit{sms\_new.asp} in \href{https://www.exploit-db.com/exploits/41074}{ExploitDB:41074}), or functions (e.g., \textit{ImageIO\_Malloc} in the exploit in \href{https://www.exploit-db.com/exploits/47970}{ExploitDB:47970} in Fig. \ref{fig:exp_exploitdb}).
As seen in these examples, the names of vulnerable components vary in morphological forms.

Vulnerability type refers to abstract software weakness of a vulnerability, 
Entries in Common Weakness Enumeration (CWE)~\cite{cwe}, for example, \href{https://cwe.mitre.org/data/definitions/119.html}{\textit{Buffer Overflow} (CWE-119)}, \href{https://cwe.mitre.org/data/definitions/79.html}{\textit{Cross-Site Scripting} (CWE-79)}, \href{https://cwe.mitre.org/data/definitions/89.html}{\textit{SQL Injection} (CWE-89)}, are commonly used to describe vulnerability type.
Similar to product names, the same vulnerability type may be described in different forms, for example, Buffer Overflow is often mentioned as overflow, Cross-Site Scripting is often written as XSS.

We consider vulnerable product/version/component and vulnerability type as four categories of software-specific entities.
We adopt machine-learning-based NER to extract these four key aspects from the exploit's title and description.
As shown in Fig.~\ref{fig:exp_exploitdb}, given natural language sentences, the NER model identifies all entities mentioned in the sentences and assign each identified entity a category.
An entity mention can be one word or several consecutive words.
Existing machine-learning based methods~\cite{dehengsaner2016, dehengicsme2016} require a gazetteer of known entity names and mention variations, which limits the deployability of these methods.
In this work, we remove this limitation by adopting the BERT-based NER.
BERT is the state-of-the-art pre-trained language model~\cite{BERT}.
One of the downstream applications BERT supports is NER.
By learning from a vast amount of natural language data, BERT can resolve mention variations, polysemy and co-references.
When applying BERT to a specific domain and a downstream task, pre-trained BERT should be fine tuned.
We manually label the title and description of the 765 exploit posts (as illustrated in Fig.~\ref{fig:exp_exploitdb}) which contain 8,717 sentences for fine-tuning BERT-based NER model (see Section~\ref{sec:dataset}).

An exploit post may mention multiple products and their versions.
For example, the \href{https://www.exploit-db.com/exploits/15935}{ExploitDB:15935} mentions not only the vulnerable product \textit{regcomp}, but also the similar product \textit{GNU libc} and the operating systems (OS) (e.g., \textit{Ubuntu} and \textit{FreeBSD}) on which the exploit has been tested.
The OS also has version information, such as \textit{Ubuntu 10.10} and \textit{FreeBSD 8.1}.
However, this OS version has nothing to do with the vulnerable product.
An exploit may mention several versions and components of a vulnerable product. 
For example, the \href{https://www.exploit-db.com/exploits/40725}{ExploitDB:40725}, states ``... Sophos Web Appliance ... Affected Version: v4.2.1.3 ... vendor has issued a fix for this vulnerability in Version
4.3 of SWA''.
Here, the vulnerable version is \textit{v4.2.1.3}, but not \textit{4.3}.
Also note that vulnerable product and vulnerable version may not always be mentioned closely.
The \href{https://www.exploit-db.com/exploits/48232}{ExploitDB:48232} states ``...the SUID executable vmware-authd is also vulnerable to this bug ... vmware-authd tries to load two libraries, libcrypto and libssl, from the incorrect directory.''
Three components vmware-authd, libcrypto and libssl are mentioned here, but only \textit{vmware-authd} is the vulnerable component.
When labeling the exploit posts, we label only vulnerable products/versions/components, but not non-vulnerable ones.
As BERT-based NER considers the global sentence context in which the entities appear, it can learn to distinguish vulnerable entities from non-vulnerable entities from the sentence context.

\subsubsection{Extracting Vendor and Attacker Type by Gazetteers}
\label{sec:vendorattackertype}

Vendor and attacker type are also software-specific entities.
For these two aspects, we do not use machine-learning-based NER for two reasons.
First, the information is often available in the exploit metadata which is easy to extract.
Second, the information may not be available in the exploit post, which has to be inferred from other sources.
Therefore, We build gazetteers to extract or infer vendor and attacker type from the exploit post and the historical vulnerability data.

For vendor, we build a product-vendor dictionary from the Common Platform Enumeration (CPE) database~\cite{CPE} of Mitre, which records the product-vendor mappings for all existing CVEs.
For a product name mentioned in the exploit, we search this dictionary to find the corresponding vendor.
If no vendor can be found, we resort to the vendor homepage metadata of the exploit.
We use the website domain name as the vendor.
If no vendor homepage is available, we use the first word of the product name as the vendor name, as the product name often starts with the vendor information (e.g., \emph{BMC Track-It!}).

An exploit must specific the attacker type metadata.
ExploitDB defines four attacker types (remote, local, webapp and DoS (Denial of Service)).
These four types overlap but are also different from the attacker types (remote, local, physical, content-dependent) CVE defines.
Therefore, we extract the exploit attacker type from the metadata, and then map it to the CVE attacker type.
Remote and local are direct mappings.
Webapp in ExploitDB is mapped to remote in CVE because the exploits use web applications as the attack medium.
DoS attack can be remote (e.g., \href{https://www.exploit-db.com/exploits/47097}{ExploitDB:47097}) or local (e.g., \href{https://www.exploit-db.com/exploits/2541}{ExploitDB:2541}).
We further search the PoC code for finding remote-indicating keywords (e.g., http, router, web, sock, ipv4/ipv6, ping, port, message).
If some remote-indicators are found, DoS is mapped to remote in CVE, otherwise local.
As the physical and context-dependent is very rare in the CVEs, we exclude them in our analysis.

\subsubsection{Extracting Root Cause, Attack Vector and Impact by BERT-QA}
Root cause, attack vector and impact are more complicated information than the six types of entities processed above.
These three aspects sometimes are a short phrase.
For example, \emph{boundary error} is a root cause, \emph{malformed TIFF image} is an attack vector, and \emph{system crash} is an impact.
However, more often than not, these three aspects are a long clause or sentence, for example, \emph{invalid format specifier used when displaying a user-supplied username} for the root cause.
Another challenge is that root cause, attack vector and impact may be mentioned several time but not in the exact same expression, as shown in Fig.~\ref{fig:exp_exploitdb}

To extract root cause, attack vector and impact, we resort to the recent advance of BERT-based Question Answering model~\cite{BERT}.
BERT QA is based on pre-trained BERT model for finding question answers in a given content scope.
The input includes a question and the scope for answering the question, and the output is the start and end word index of the answer clause.
In our application of BERT QA, we use the description of an exploit post as the scope of question answering, and we ask the model to find the answer to three what-is questions ``what is root cause?'', ``what is attack vector?'' and ``what is impact?''.
Due to the language modeling capability of BERT, BERT QA can handle the complex clauses of root cause, attack vector and impact and select the most appropriate information from the long exploit text.

We train the BERT-QA model with 3,693 question-answer pairs manually constructed from 1,231 exploit posts (see Section~\ref{sec:bertqadataset}).
We create both positive and negative questions for which the answers can or cannot be found in the given exploit posts.
The negative questions allows the model to learn when it cannot find answers in the scope.
This characteristic is important for extracting root cause, attack vector and impact, because not all exploit posts describe these three aspects.
Without the ability to handle negative questions, the model will be forced to extract some irrelevant content as the answers.

\subsection{Composing CVE Description from Extracted Aspects}
\label{sec:compose}

After extracting the needed aspects from an exploit, we select the most informative aspect instances and use them to construct a CVE description according to the CVE templates.

\subsubsection{Selecting Informative Aspect Instances}
For vendor, attacker type, root cause, attack vector and impact, our approach extracts a single instance from an exploit post.
For vulnerable version and component, they can have multiple instances in an exploit post.
As the exploit may actually involve multiple vulnerable versions and components, we keep all unique versions and components being extracted.
Our approach may also extract multiple product names and vulnerability types from an exploit post (see Fig.~\ref{fig:exp_exploitdb}).
We develop heuristics to select the most informative product name and vulnerability type.
First, we prefer longer names, for example, \emph{Microsoft Window 10} versus \emph{Win10}, \emph{cross-site scripting} versus \emph{XSS}.
Second, we prefer the names being mentioned more frequently in the exploit post, as the term frequency indicates the importance of the product name and vulnerability type in the exploit.

\subsubsection{Filling in the CVE Template}

Finally we compose the CVE description by filling the extracted aspects into the CVE templates~\cite{CVEtemplate} recommended by Mitre.
There are two templates: ``[VULTYPE] in [COMPONENT] in [VENDOR][PRODUCT][VERSION] allows [ATTACKER] to [IMPACT] via [VECTOR]'', and ``[COMPONENT] in [VENDOR][PRODUCT][VERSION] [ROOT CAUSE], which allows [ATTACKER to [IMPACT] via [VECTOR]''.
If vulnerability type is extracted, we use the first template.
If no vulnerability type is extracted but root cause is extracted, we use the second template.
If neither vulnerability type nor root cause is extracted, we use the first template as the sentence is more natural.
When certain aspect is not extracted, the corresponding placeholder is left empty.
We adjust the sentence form to keep the composed CVE description natural.
For example, when attacker type is not extracted, we adjust ``allow [ATTACKER] to [IMPACT]'' to ``cause [IMPACT]'' which is a common expression found in CVE descriptions.

\section{Accuracy of Aspect Extraction from Exploits}
\label{sec:eva}

First, we evaluate the accuracy of the NER method for extracting vulnerable products/versions/components and vulnerability types, and the BERT-QA method for extracting root causes, attack vectors and impacts.

\subsection{Accuracy of BERT-based NER}

\subsubsection{Datasets}
\label{sec:dataset}

We manually label 765 exploit posts for evaluating the NER methods.
These exploit posts contain 8,717 sentences with 102,262 word tokens.
Among these sentences, 2,445 (28\%) sentences contain at least one of the four aspects to be extracted.
In total, we label 2,007, 1,061, 1,430 and 525 instances of vulnerable products, versions, components and vulnerability types, respectively.
The scale of this ExploitDB corpus is on par with other general NER datasets~\cite{Hermann@2015} and software-specific NER datasets~\cite{dehengsaner2016, dehengicsme2016, masuyutse2019}.

We also use a corpus of CVE descriptions~\cite{CVE} for training the NER model.
This corpus contains 90,313 CVE descriptions like the example in Fig.~\ref{fig:exp_nvd}, with in total 87,384, 50,263, 30,157 and 33,267 instances of vulnerable products, versions, components and vulnerable types, respectively. 
The goal is to investigate whether learning from the target CVE descriptions we aim to compose may help the model extract the needed information from the source exploits.
This CVE-description corpus is larger than our ExploitDB corpus, but it is not fully manually labelled.
Instead, it uses a set of regular expressions to label vulnerable products, versions, components and vulnerability types in the CVE descriptions, and then adopts a statistically sampling method~\cite{singh2013elements} to validate the identified entities.
The validation confirms the high accuracy ($>$97\%) of the identified entities.
This type of semi-automatic labelling method is often used in the NLP community to create large-scale (but maybe noisy) corpora~\cite{torr2017autobank,rehbein2017detecting,kshirsagar2015frame}. 

\subsubsection{Baselines}
We use two baselines to compare with our BERT-based NER.
The first baseline is the Long Short Term Memory (LSTM) based NER model~\cite{inconsistance@usenix, masuyutse2019}.
Similar to our BERT-based method, this LSTM-based NER model does not require entity-name gazetteer either. 
Instead, it uses multi-level feature learning to capture character, word and sentence-context features for identifying software-specific entities (e.g., products, libraries, concepts, APIs) in software text.
For the second baseline, we try our best to summarize entity-extraction rules during the manual labeling of the ExploitDB corpus.

\subsubsection{Evaluation Methods}
Model training and testing are done for each aspect independently.
For each aspect, we randomly split the sentences in the ExploitDB corpus into 80\% and 20\% for model training and testing respectively.
We use three training-dataset settings: CVE-only, ExploitDB-only, and CVE$\rightarrow$ExploitDB (i.e., train the NER model with the CVE corpus first and then fine-tune the model with the ExploitDB corpus).
The same ExploitDB testing data is used to evaluate the models and baselines obtained in different dataset settings.
We perform 5-fold cross-validation and report the average performance.
We use precision, recall and F1-score to evaluate the entity-extraction performance, which are commonly used for evaluating NER methods~\cite{dehengsaner2016,dehengicsme2016,masuyutse2019}.

\subsubsection{Results}

\begin{table}[tb]
	\centering
	\caption{Comparison between NER Models and Training Datasets}
	\vspace{-3mm}
	\label{tab:acc_ner}
	\begin{tabular}{|c|c|c|c|c|}
		\hline
		Model&Dataset&Precision&Recall&F1 Score\\
		\hline
		\multirow{3}{*}{BERT}&CVE-only&0.52& 0.08&0.14\\
		\cline{2-5}
		&ExploitDB-only&0.74&\textbf{0.79}&\textbf{0.76}\\
		\cline{2-5}
		&CVE$\rightarrow$ExploitDB&0.74&0.78&\textbf{0.76}\\
		\hline
		\multirow{3}{*}{LSTM}&CVE-only&0.43&0.21&0.28\\
		\cline{2-5}
		&ExploitDB-only&0.76&0.67&0.71\\
		\cline{2-5}
		&CVE$\rightarrow$ExploitDB&\textbf{0.84}&0.64&0.73\\
		\hline
	\end{tabular}
	\vspace{-1mm}
\end{table}

\begin{table}[tb]
	\centering
	\caption{BERT-based NER versus Rule-based NER}
	\vspace{-3mm}
	\label{tab:acc_ner_each}
	\begin{tabular}{|c|c|c|c|c|c|c|}
		\hline
		&\multicolumn{2}{c|}{Precision}&\multicolumn{2}{c|}{Recall}&\multicolumn{2}{c|}{F1 Score}\\
		\cline{2-7}
		&BERT&Rule&BERT&Rule&BERT&Rule\\
		\hline
		Product&\textbf{0.76}&0.55&\textbf{0.82}&0.59&\textbf{0.79}&0.57\\
		\hline
		Version&\textbf{0.77}&0.60&\textbf{0.80}&0.66&\textbf{0.78}&0.63\\
		\hline
		Vultype&0.70&\textbf{0.78}&\textbf{0.78}&0.43&\textbf{0.74}&0.55\\
		\hline
		VulComp&0.72&\textbf{0.91}&\textbf{0.70}&0.54&\textbf{0.71}&0.68\\
		\hline
	\end{tabular}
	\vspace{-6mm}
\end{table}

Table~\ref{tab:acc_ner} shows the metrics of the four types of entities as a whole for the six combinations of model and training-dataset settings.
To our surprise, both BERT- and LSTM-based NER models perform poorly on the CVE-only corpus, especially for recalls on vulnerability type and vulnerable component.
This can be attributed to the lexical gap between the CVE descriptions and the ExploitDB posts.
For example, CVE often describes vulnerable components at the package/module granularity, while ExploitDB tends to describe specific vulnerable files or functions.
Therefore, the features learned from the CVE-only corpus do not generalize well to the ExploitDB corpus.

Fine-tuning the CVE-trained model with the ExploitDB corpus can significantly boost the performance.
However, this CVE-train-ExploitDB-finetuning setting does not seem necessary, as simply training the NER model with the ExploitDB-only corpus achieves almost the same performance as the complex CVE-train-ExploitDB-finetuning setting.
When training with the ExploitDB corpus, the LSTM-based model achieves better precision, while the BERT-based model achieves better recall.
The BERT-based model achieves more balanced precision and recall, and thus the better F1-score overall.

Table~\ref{tab:acc_ner_each} compares the performance of the BERT-based NER trained with the ExploitdB-only corpus with the performance of the rule-based NER.
We show the performance metrics for the four types of entities separately.
The BERT-based NER achieves much higher recalls and F1-scores for all four types of entities, compared with the rule-based NER.
The precisions of the rule-based NER are also much worse than those of the BERT-based NER for product and version.
Our results echo the findings of existing software-specific NER studies~\cite{dehengsaner2016, dehengicsme2016, inconsistance@usenix}, because it is difficult to develop a comprehensive set of rules to handle all mention variations and unseen names.
However, the rule-based NER may extract entities with better precision when there are a small number of frequently-mentioned entities (e.g., vulnerability types) or the entities have distinct orthographic features (e.g., brackets, hyphenation, capitalization in components).

Comparing the performance of BERT-based NER for the four types of entities in Table~\ref{tab:acc_ner_each}, we see that vulnerable component is the most challenging entities to recognize.
This is consistent with the findings in~\cite{dehengsaner2016}, which shows that API entities are more challenging to recognize correctly than other types of entities like products and concepts.
Recognizing vulnerability types has relatively low F1 due to low precision.
A common mistake BERT-based NER makes is to recognize single word (e.g., overflow, service) as vulnerability type.
As shown in Fig.~\ref{fig:exp_exploitdb}, people sometimes do use single word to describe vulnerability type (e.g., the overflow happens ...).
But more often than not, this is not the case.

\vspace{1mm}
\noindent\fbox{\begin{minipage}{8.6cm} \emph{BERT-based NER outperforms rule-based NER. It also outperforms LSTM-based NER, and achieves more balanced precision and recall. It is sufficient to train BERT-based NER model with only the ExploitDB corpus. Among the four aspects, products and versions are easier to recognize than vulnerable components and vulnerability types} \end{minipage}}\\

\subsection{Accuracy of BERT QA}

\subsubsection{Datasets}
\label{sec:bertqadataset}

We manually label 1,231 ExploitDB posts for evaluating the extraction of root causes, attack vectors and impacts by BERT QA.
We need to label more data than the NER corpus because root causes and attack vectors are mentioned less frequently than vulnerable products, versions and components.
The 1,231 ExploitDB posts contain 4,300 sentences, among which 1,877 (43\%) sentences contain at least one of the three needed aspects.
In total, we label 318 root causes, 653 attack vectors and 1,454 impacts in these 1,877 sentences.
Based on the labelling results, we create question-answer pairs for the 1,231 exploit posts, including 318, 653 and 1,454 positive question-answer pairs for root cause, attack vector and impact respectively, and the same amount of negative question-answer pairs for the corresponding aspects.

In a similar way, based on the labels of root causes, attack vectors and impacts in the corpus of 90,313 CVE descriptions~\cite{CVE}, we create 11,757, 79,668 and 91,893 positive question-answer pairs for root cause, attack vector and impact, respectively, and the same amount of negative question-answer pairs for the three aspects.
We also train the BERT-QA model with the general question-answering dataset SQuAD v2.0.
It contains 97,279 positive question-answer pairs from 536 English Wikipedia articles, and 53,775 negative question-answer pairs.
Using the CVE dataset and the SQuAD dataset allows us to investigate whether the question-answer features in the general text and the CVE descriptions may benefit the BERT-QA's capability of finding answers in the exploit text. 

\subsubsection{Baseline}

During the manual labeling of the ExploitDB posts, we try our best to develop a rule-based extraction baseline.
These rules look for specific keywords or sentence patterns which often indicate root causes (e.g., due to, failure to do), attack vectors (e.g., malformed, specially crafted), or impacts (e.g., lead to, result in, cause the application to, attacker can, allow attacker to).
We extract the clauses enclosing the matched keywords or patterns as the answer candidates to the root-cause, attack-vector or impact question.
If there are multiple answer candidates, we choose the longest candidate because the longest candidate would involve more complete information, for example, \emph{allows remote attackers to crash the application, denying further service to legitimate users} versus \emph{allows remote attackers to cause daemon hang}.

\begin{table}[tb]
	\centering
	\caption{Results of BERT-QA Trained with Different Datasets}
	\label{tab:acc_qa}
	\begin{tabular}{|c|c|c|c|c|c|c|}
		\hline
		&\multicolumn{3}{c|}{Exact-Match Score}&\multicolumn{3}{c|}{F1 Score}\\
		\cline{2-7}
		&Over&Posi&Nega&Over&Posi&Nega\\
		&all&tive&tive&all&tive&tive\\
		\hline
		SQuAD-only&0.44&0.07&0.81&0.50&0.19&0.81\\
		\hline
		CVE-only&0.50&0.01&\textbf{0.99}&0.51&0.01&\textbf{0.99}\\
		\hline
		ExploitDB-only&\textbf{0.77}&\textbf{0.58}&0.95&\textbf{0.82}&\textbf{0.68}&0.95\\
		\hline
		SQuAD$\rightarrow$CVE&0.50&0.01&\textbf{0.99}&0.50&0.01&\textbf{0.99}\\
		\hline
		SQuAD$\rightarrow$&\multirow{2}{*}{0.76}&\multirow{2}{*}{\textbf{0.58}}&\multirow{2}{*}{0.94}&\multirow{2}{*}{\textbf{0.82}}&\multirow{2}{*}{\textbf{0.68}}&\multirow{2}{*}{0.94}\\
		ExploitDB&&&&&&\\
		\hline
		CVE$\rightarrow$&\multirow{2}{*}{0.75}&\multirow{2}{*}{0.56}&\multirow{2}{*}{0.94}&\multirow{2}{*}{\textbf{0.82}}&\multirow{2}{*}{\textbf{0.68}}&\multirow{2}{*}{0.94}\\
		ExploitDB&&&&&&\\
		\hline
		SQuAD$\rightarrow$CVE&\multirow{2}{*}{0.75}&\multirow{2}{*}{0.55}&\multirow{2}{*}{0.94}&\multirow{2}{*}{0.81}&\multirow{2}{*}{\textbf{0.68}}&\multirow{2}{*}{0.94}\\
		$\rightarrow$ExploitDB&&&&&&\\							
		\hline
	\end{tabular}
	\vspace{-1mm}
\end{table}


\begin{table}
	\centering
	\renewcommand{\tabcolsep}{1.6pt}
	\caption{BERT-QA versus Rule-based Extraction}
	\label{tab:acc_qa_each}
	\begin{tabular}{|c|c|c|c|c|c|c|c|c|c|c|c|c|}
		\hline
		&\multicolumn{6}{c|}{Exact-Match Score}&\multicolumn{6}{c|}{F1 Score}\\
		\cline{2-13}
		&\multicolumn{3}{c|}{BERT-QA}&\multicolumn{3}{c|}{Rule-based}&\multicolumn{3}{c|}{BERT-QA}&\multicolumn{3}{c|}{Rule-based}\\
		\cline{2-13}
		&Over&Posi&Negi&Over&Posi&Negi&Over&Posi&Negi&Over&Posi&Negi\\
		&all&tive&tive&all&tive&tive&all&tive&tive&all&tive&tive\\
		\hline
		Root&\multirow{2}{*}{\textbf{0.79}}&\multirow{2}{*}{\textbf{0.59}}&\multirow{2}{*}{0.98}&\multirow{2}{*}{0.57}&\multirow{2}{*}{0.14}&\multirow{2}{*}{\textbf{0.99}}&\multirow{2}{*}{\textbf{0.87}}&\multirow{2}{*}{\textbf{0.75}}&\multirow{2}{*}{0.98}&\multirow{2}{*}{0.58}&\multirow{2}{*}{0.16}&\multirow{2}{*}{\textbf{0.99}}\\
		Cause&&&&&&&&&&&&\\
		\hline
		Vector&\textbf{0.67}&\textbf{0.43}&0.91&0.58&0.15&\textbf{0.99}&\textbf{0.77}&\textbf{0.62}&0.91&0.63&0.26&\textbf{0.99}\\
		\hline
		Impact&\textbf{0.82}&\textbf{0.74}&0.90&0.78&0.58&\textbf{0.95}&\textbf{0.87}&\textbf{0.83}&0.90&0.82&0.68&\textbf{0.95}\\
		\hline
	\end{tabular}
	\vspace{-4mm}
\end{table}

\subsubsection{Evaluation Method}
Model training and testing is done for each aspect independently.
For each aspect, we randomly split the question-answer pairs in the ExploitDB dataset into 80\% and 20\% for model training and testing respectively.
In addition to the ExploitDB question-answer pairs, we also train the model with the SQuAD  dataset and the CVE dataset.
This gives us six training dataset settings: SQuAD-only, CVE-only, ExploitDB-only, SQuAD$\rightarrow$CVE, SQuAD$\rightarrow$ExploitDB, CVE$\rightarrow$ExploitDB, SQuAD$\rightarrow$CVE$\rightarrow$ExploitDB.
X$\rightarrow$Y means training with X first and then fine-tuning with Y.
The same ExploitDB testing data is used to evaluate the models and baselines obtained in different dataset settings.
We perform 5-fold cross-validation and report the average performance.

We adopt the metrics in~\cite{BERT} to evaluate the answer quality of a QA system.
For both positive and negative questions, we compute two metrics: exact-matching and F1-score.
For a positive question, exact-matching score is 1 if the answer clause returned by the QA system matches exactly with the ground-truth answer clause.
Otherwise, exact-matching score is 0.
F1-score allows partial matching. 
It is the percentage of matched words between the returned answer clause and the ground-truth answer clause.
For a negative question, both exact-matching and F1-score is 1 if the QA system correctly returns no answers for the question, otherwise 0.
By summing the scores of positive and negative questions, we obtain the overall scores.
We normalize the scores by dividing them by the number of positive, negative and all questions.

\subsubsection{Results}

As shown in Table \ref{tab:acc_qa}, training the BERT-QA model without the ExploitDB data (e.g., SQuAD, CVE and SQuAD$\rightarrow$CVE) cannot produce satisfactory answer quality.
The model trained with only the general text (SQuAD) has the worst overall performance.
Training the model with CVE descriptions can improve the overall performance.
This overall improvement comes from almost perfect performance (0.99) for negative questions.
That is, for 99\% of the exploit posts without root causes, attack vectors or impacts, BERT-QA correctly returns no answers for absent aspects.
However, for the exploits with root causes, attack vectors or impacts (i.e., positive questions), the model trained with CVE data completely fails to retrieve the aspects present in the exploits (0.01 exact-matching score and 0.01 F1-score).
This is because the model does not learn well from the CVE descriptions for extracting information from the exploit posts.
As such, it either returns no answers or just some meaningless incomplete phrases (e.g., to come out, a vulnerability)

As long as the model is trained with the ExploitDB question-answer pairs, it does not matter that much whether the model is pre-trained with SQuAD and/or CVE.
For the ExploitDB-only setting, the model achieves 0.77 overall exact-matching score and 0.82 overall F1-score.
For negative questions, the model has 0.14 higher exact-matching and F1-scores than the model trained with SQuAD, and has slightly (0.04)  lower scores than the model trained with CVE.
For the positive questions, the model trained with ExploitDB achieves 0.58 exact-matching scores.
That is, the model returns the exact ground-truth aspects for 58\% of the exploit posts.
For partially matched aspects, the F1-score is 0.68.

Table \ref{tab:acc_qa_each} compares the performance of the BERT-QA model trained with the ExploitDB-only data with the performance of the rule-based extraction.
We show the performance metrics for the three aspects separately.
For negative questions without answers, rule-based extraction has slightly higher exact-matching and F1 scores than BERT-QA.
Rule-based extraction tends to returns no answers due to the incompleteness of rules, while BERT-QA may extract some irrelevant information for negative questions.
For positive questions with answers, BERT-QA achieves significantly higher exact-matching and F1-scores than the rule-base extraction.
Rule-based extraction fails to extract aspect information from many exploits, because it is hard to have a comprehensive set of rules to cover all mention situations.
Rules can better extract impacts than extracting root causes and attack vectors.
This is because different vulnerabilities may cause similar impacts, but root cause and attack vector are often specific to a vulnerable component and exploit method.

BERT-QA has the best scores for positive questions on impact, while it has the close scores for negative questions on the three aspects.
The overall scores for impact and root cause are very close.
BERT-QA sometimes extracts intermediate impacts (e.g., create files on the file system in \href{https://www.exploit-db.com/exploits/4148}{ExploitDB:4148}) as final impacts.
For root cause, BERT-QA often extracts only partial root cause clauses, which results in lower extract-matching score (0.59) than F1 score (0.75).
BERRT-QA performs relatively poorer for positive questions on attack vector, especially for extracting exact-matching attack vector (0.43).
Compared with root cause and the impact, attacker vector is often more complex.
In many cases, attack vector refers to specific implementation of an exploit method (e.g. redirect kernel code execution to 0x40404040 in \href{https://www.exploit-db.com/exploits/7460}{ExploitDB:7460}, pass a string over 2068 in \href{https://www.exploit-db.com/exploits/39629}{ExploitDB:39629}), which requires deep software and security knowledge to identify.

\vspace{1mm}
\noindent\fbox{\begin{minipage}{8.6cm} \emph{Although rule-based extraction can extract some impact information, it fails to retrieve most of root causes and attack vectors. In contrast, the BERT-QA trained with the ExploitDB question-answer pairs can effectively extract impacts and root causes in the exploit posts. Accurately extracting attack vector is more challenging as it requires a deeper understanding of the exploit methods. Pre-training the model with general text and CVE descriptions does not affect the model performance.} \end{minipage}}\\

\section{Quality and Usefulness of Composed CVEs}
\label{sec:use}

Using our method, we compose 41,883 CVE descriptions for 41,883 exploits as of December 31 2019, including 25,670 exploits with CVE IDs and 16,213 exploits with CVE IDs.
We exclude 1,764 labelled exploits (765 for BERT-NER and 1,231 for BERT-QA with 232 overlaps) used for training or developing the aspect-extraction methods.
These composed CVE descriptions involve 41,413 product names, 40,120 versions, 24,624 vulnerable components, 40,372 vulnerability types, 10,657 root causes, 12,456 attack vectors and 21,370 impacts
In this section, we investigate the quality and usefulness of the composed CVE descriptions.

\subsection{Correctness of Composed CVE Aspects}

We first investigate the correctness of individual aspects in the composed CVE descriptions. 

\subsubsection{Evaluation Method}
Considering the large number of exploits, we use a statistical sampling method~\cite{singh2013elements} to examine 384 extracted instances of each CVE aspect respectively, 
This allow us to estimate the accuracy of our method on extracting each aspect with error margin 0.05 at 95\% confidence level.
If the exploit has a corresponding CVE, we consider the description of that CVE as the reference description.
We examine an extracted aspect in the composed CVE description against the corresponding aspect in the reference CVE description.
As these two aspects may have lexical gap, we determine the correctness by matching aspect semantics.
If the composed description contains some aspects not present in the reference description, we determine the correctness of the extracted aspects against the exploit post content.
If the exploit does not have a corresponding CVE, we also determine the correctness of the extracted aspect against the exploit content.
Two authors annotate the correctness of the sampled data instances independently. 
Afterwards, they discuss to resolve the disagreements.
We compute the accuracy based on the final annotation consensus.
We compute Cohen's Kappa to evaluate the inter-rater agreement.

\subsubsection{Results}
The Cohen's Kappa for the nine aspects ranges from 0.63 to 0.76, which indicates substantial agreement between the two annotators.
The accuracy for the sampled products, versions, components, vulnerability types, root causes, attack vectors and impacts is 96.59\%, 92.39\%, 97.65\%, 93.67\%, 96.24\%, 94.18\%, and 92.80\%, respectively.
The erroneous aspects are mainly due to the errors made by the NER and QA models.
The accuracy of the 384 sampled vendors and attacker types is 80\% and 94\% respectively.
The vendor accuracy is relatively low for two reasons.
First, some products (e.g., QiHang Media, 13Plugins) cannot be found in the CPE database.
Second, when there is no explicit product-vendor information, inferring the vendor from the vendor homepage metadata or the product name is not always reliable.
For example, the site Sourcecodester is a open source platform which is not the real vendor of the products hosted on the platform, or Java CMM's vendor is Oracle rather than Java.

\vspace{1mm}
\noindent\fbox{\begin{minipage}{8.6cm} \emph{The composed CVE descriptions contain highly accurate aspects extracted from the exploits.} \end{minipage}}\\

\subsection{Similarity of Composed and Reference CVE Descriptions}


If the exploit has a corresponding CVE, we consider the description of that CVE as the reference description.
Next, we evaluate the overall similarity of the composed CVE descriptions and the reference CVE descriptions.

\subsubsection{Experiment Setup}
Our approach is a white-box text summarization method, which explicitly extracts different CVE aspects and then compose them.
We compare our approach with three black-box text summarization methods: PreSumm \cite{liu@emnlp2019}, SummaRuNNer \cite{Nallapati@2017}, and BanditSum \cite{dong@2018acl}.
All the three methods use deep learning models for generating summaries for long texts. 
We train these three deep learning baselines using the title and main context of an exploit as input and the description of the corresponding CVE as the output.

There are 27,230 exploits with corresponding CVEs.
For the three deep learning baselines, we use 80\% of the exploits as the training data, and the rest 20\% as the testing data.
We use our method to compose the CVE descriptions for the testing exploits.
We adopt ROUGE-1, ROUGE-2 and ROUGE-L which are the metrics for evaluating text summarization methods \cite{Rouge}.
ROUGE-1 (or 2) refer to the overlap of unigram (or bigrams) between the composed and reference descriptions.
ROUGE-L measures the longest in-sequence common n-grams using Longest Common Subsequence algorithm.
We perform 5-fold cross validation and report the average performance.


\subsubsection{Result Analysis}
\label{sec:rouge}

\begin{table}[tb]
	\centering
	\caption{Similarity of Composed and Reference CVE Descriptions}
	\vspace{-3mm}
	\label{tab:rouge}
	\begin{tabular}{|c|c|c|c|}
		\hline
		&ROUGE-1&ROUGE-2&ROUGE-L\\
		\hline
		SummaRuNNer&0.28&0.12&0.24\\
		\hline
		BanditSum&0.27&0.11&0.24\\
		\hline
		PreSumm&0.33&0.14&0.28\\
		\hline
		Our Method&\textbf{0.41}&\textbf{0.16}&\textbf{0.38}\\
		\hline
	\end{tabular}
	\vspace{-6mm}
\end{table}

\begin{table*}[tb]
	\centering
	\caption{Examples of Reference and Composed CVE Descriptions}
	\label{tab:composedreferenceexamples}
	\vspace{-2mm}
	\begin{tabular}{|l|l|}
		\hline
		\multicolumn{1}{|c|}{Reference CVE Descriptions}&\multicolumn{1}{c|}{Composed CVE Descriptions}\\
		\hline
		\textbf{\href{https://nvd.nist.gov/vuln/detail/CVE-2010-4557}{CVE-2010-4557}:} \textcolor{mygreen}{Buffer overflow} in the \textcolor{mylightorange}{lm\_tcp} service in \textcolor{mypurple}{Invensys Wonderware}&\textbf{\href{https://www.exploit-db.com/exploits/15707}{ExploitDB:15707}:} \textcolor{mygreen}{buffer overflow} in \textcolor{mylightorange}{lm\_tcp} in \textcolor{mypurple}{invensys's} \\ 
		\textcolor{mypurple}{InBatch} \textcolor{mybrown}{8.1} and \textcolor{mybrown}{9.0}, as used in \textcolor{mypurple}{Invensys Foxboro I/A Series Batch} \textcolor{mybrown}{8.1} and& \textcolor{mypurple}{WonderWare InBatch} \textcolor{mybrown}{9.0sp1} allows remote attacker to cause \textcolor{myblue}{denial}\\
		possibly other products, allows remote attackers 	to cause \textcolor{myblue}{a denial of service}  & \textcolor{myblue}{of service} via \textcolor{mypink}{writing a 16bit 0x0000 in an arbitrary memory location}\\
		\textcolor{myblue}{(crash)} and possibly \textcolor{myblue}{execute arbitrary code} via \textcolor{mypink}{a crafted request to port 9001}&\\

		\hline
		\textbf{\href{https://nvd.nist.gov/vuln/detail/CVE-2013-6767}{CVE-2013-6767}:} \textcolor{mygreen}{Stack-based buffer overflow} in \textcolor{mylightorange}{pepoly.dll} in \textcolor{mypurple}{Quick Heal}&\textbf{\href{https://www.exploit-db.com/exploits/30374}{ExploitDB:30374}:} \textcolor{mygreen}{stack buffer overflow} in \textcolor{mylightorange}{pepoly.dll} in \textcolor{mypurple}{QuickHeal's}\\ 
		 \textcolor{mypurple}{AntiVirus Pro} \textcolor{mybrown}{7.0.0.1} allows local users to \textcolor{myblue}{execute arbitrary code} or cause \textcolor{myblue}{a}& \textcolor{mypurple}{AntiVirus Pro} \textcolor{mybrown}{7.0.0.1} and \textcolor{mybrown}{7.0.0.1 (b2.0.0.1)} allows local attacker to \\
		 \textcolor{myblue}{denial of service (process crash)} via \textcolor{mypink}{a long *.text value in a PE file}& cause \textcolor{myblue}{stack overflow} via \textcolor{mypink}{a manipulated import of a malicious pe file}\\
		 
		 \hline
		 \textbf{\href{https://nvd.nist.gov/vuln/detail/CVE-2016-8580}{CVE-2016-8580}:} \textcolor{mygreen}{PHP object injection vulnerabilities} exist in \textcolor{mylightorange}{multiple widget} &\textbf{\href{https://www.exploit-db.com/exploits/40682}{ExploitDB:40682}:} \textcolor{mygreen}{php object injection} in \textcolor{mylightorange}{image.php} in \textcolor{mypurple}{alienvault's}\\ 
		 \textcolor{mylightorange}{files} in \textcolor{mypurple}{AlienVault OSSIM} and \textcolor{mypurple}{USM} \textcolor{mybrown}{before 5.3.2}. These vulnerabilities allow& \textcolor{mypurple}{OSSIM/USM} \textcolor{mybrown}{5.3.1} allows remote attacker to \textcolor{myblue}{gain code execution} via \\
		  \textcolor{myblue}{arbitrary PHP code execution} via \textcolor{mypink}{magic methods in included classes}&\textcolor{mypink}{sending a serialized php object to one of the vulnerable pages}\\
		 
		 \hline
		 \textbf{\href{https://nvd.nist.gov/vuln/detail/CVE-2018-20580}{CVE-2018-20580}:}\textcolor{mylightorange}{The WSDL import functionality} in \textcolor{mypurple}{SmartBear  ReadyAPI}&\textbf{\href{https://www.exploit-db.com/exploits/46796}{ExploitDB:46796}:} \textcolor{mygreen}{code execution} in \textcolor{mypurple}{smartbear's ReadyAPI} \textcolor{mybrown}{2.5.0} \\ 
		 \textcolor{mybrown}{2.5.0} and \textcolor{mybrown}{2.6.0} allows remote attackers to \textcolor{myblue}{execute arbitrary Java code} via & and \textcolor{mybrown}{2.6.0} allows remote attacker to \textcolor{myblue}{cause code execution} via \textcolor{mypink}{opening}\\
		  \textcolor{mypink}{a crafted request parameter in a WSDL file}& \textcolor{mypink}{a soap project and import wsdl files}\\
		 \hline
	\end{tabular}
	\vspace{-6mm}
\end{table*}

Table \ref{tab:rouge} shows the ROUGE results.
The three deep learning baselines have very similar performance, about 0.3 for ROUGE-1, 0.12 for ROUGE-2 and about 0.28 for ROUGE-L.
These ROUGH metrics actually are on par with the performance of these models on other text corpus.
Our method outperforms the three baselines in all three ROUGE metrics, especially for ROUGE-1 (0.41) and ROUGH-L (0.38).
Table \ref{tab:composedreferenceexamples} shows some examples of composed CVE descriptions versus reference CVE descriptions.
We see that although the composed and reference CVE descriptions have lexical gaps, the composed descriptions largely capture the important vulnerability aspects mentioned in the reference descriptions.

\vspace{1mm}
\noindent\fbox{\begin{minipage}{8.6cm} \emph{Our white-box text summarization method outperforms the deep learning based black-box methods for generating concise CVE descriptions from detailed exploit contents.} \end{minipage}}\\

\section{Discussion}
\label{sec:discussion}

\subsection{Request new CVEs using the Composed CVE Descriptions}
A direct downstream application of our approach is to request new CVEs for those exploits without CVEs using the composed CVE descriptions.
We select 60 exploits without CVEs, for which our approach composes high-quality CVE descriptions.
After making minor edits to correct grammatical errors in the composed CVE descriptions, we submit them to the Mitre for requesting new CVE IDs for the 60 exploits without CVEs.
Till the submission of this paper, we received 19 replies from the Mitre, which approved 5 of the submitted new CVE requests (\href{https://nvd.nist.gov/vuln/detail/CVE-2019-19489}{CVE-2019-19489}, \href{https://nvd.nist.gov/vuln/detail/CVE-2019-19490}{CVE-2019-19490}, \href{https://nvd.nist.gov/vuln/detail/CVE-2019-19491}{CVE-2019-19491}, \href{https://nvd.nist.gov/vuln/detail/CVE-2019-19492}{CVE-2019-19492}, \href{https://nvd.nist.gov/vuln/detail/CVE-2019-20085}{CVE-2019-20085}).

The Mitre provided the reasons for some rejected requests.
For example, the reply for the \href{https://www.exploit-db.com/exploits/47775}{ExploitDB:47775}) explains ``Mitre does not accept local buffer overflow caused by long inputs''.
However, many similar vulnerabilities (e.g., \href{https://cve.mitre.org/cgi-bin/cvename.cgi?name=2017-7720}{CVE-2017-7720} and \href{https://cve.mitre.org/cgi-bin/cvename.cgi?name=2017-8367}{CVE-2017-8367}) like the \href{https://www.exploit-db.com/exploits/47775}{ExploitDB:47775}) has been accepted.
A recent study~\cite{Chen@2018} summarizes many reasons (e.g., duplication submission, wrongly identified vulnerability, non-replicable vulnerability) that could cause new CVE request being rejected.
We believe at least one of the rejections could be due to duplicate submission.
The Mitre rejects our CVE request for the \href{https://www.exploit-db.com/exploits/47985}{ExploitDB:47985}, but an exact same \href{https://nvd.nist.gov/vuln/detail/CVE-2020-8641}{CVE-2020-8641}as what we submitted has been created.

The CVE request process is still a mystery for us.
For 3 requests, the Mitre says they are out of their scope and asks us to submit to specific vendor-and-project CNAs.
We conjecture that this could also be the reason for the rejections that the Mitre did not explain why and for those requests that we did not hear back from the Mitre.

\subsection{Ensure Consistency in Vulnerability Data}
\label{sec:augment}
The ExploitDB posts are contributed by the crowd.
As shown in Fig.~\ref{fig:exp_exploitdb}, the post content is very informal.
The exploit submitters may miss some important aspect(s) in the description.
Among the exploits we examine, only about 60\% mention vulnerable components, about half mention impacts, and about 30\% mention root causes or attack vectors.
Furthermore, the same aspects may be described in variant ways, for example, \textit{tiff image} versus \textit{malformed tiff image}, \textit{buffer overflow} versus \textit{overflow}, \textit{crash with an access violation} versus \textit{crash with a memory violation} in the Fig.~\ref{fig:exp_exploitdb} example.
Based on the CVE aspects our approach extracts from the exploit posts, we could develop a janitor bot which could suggest the submitters to add the missing aspects or normalize inconsistent mentions.

In our study of the ExploitDB exploits with CVEs, we find inconsistencies are prevalent between the corresponding exploits and CVEs.
For example, the \href{https://www.exploit-db.com/exploits/40095}{ExploitDB:40095} explains the attack vector is ``a specially crafted PDF file containing an invalid .ttf font'', but the corresponding \href{https://nvd.nist.gov/vuln/detail/CVE-2016-4205}{CVE-2016-4205} does not have this attack vector information.
As another example, the \href{https://www.exploit-db.com/exploits/42386}{ExploitDB:42386} mentions Buffer Overflow as the vulnerability type, but the corresponding \href{https://nvd.nist.gov/vuln/detail/CVE-2016-2226}{CVE-2016-2226} says Integer Overflow is the vulnerability type.
Buffer Overflow (CWE-119) and CWE of Integer Overflow (CWE-190) have a can-follow relation.
In Table~\ref{tab:composedreferenceexamples}, we can observe some other inconsistencies.
We could contrast the CVE aspects our approach extracts from the exploits against those of the corresponding CVEs to identify and reconcile the inconsistencies in between.

\section{Related Work}
\label{sec:relate}



CVE~\cite{CVE} is the most influential vulnerability database.
It is the foundation for various security analysis~\cite{Felsch@usenix2018,Wang@usenix2017,Ge@ieee2016,Goktas@ieee2018, Kim@ieee2017, Cozzi@ieee2018, Zhu@ieee2018, Martinelli@acccs2017, You@ssp2019, Huang@issp2019, Wang@ieee2017, Chen@usenix2017}
In addition to the vendor-and-project CNAs~\cite{CVERequest}, individuals can also request CVEs for the vulnerabilities they discover.
They can either report the vulnerabilities directly to the vendors or Mitre, or they may  post their discoveries on public website, such as ExploitDB~\cite{ExploitDB}, SecurityFocus~\cite{Securityfocus}, which often become references (e.g., initial announcement) for official CVEs.
Our study shows that ExploitDB is one of the most referenced crowdsourcing website by CVEs.

Beside CVE metadata (e.g., CVSS score, reference), CVE description is an important information for vulnerability analysis.
For example,  studies~\cite{hanzuobingicsme2018severityprediction,gongxiiceccs2019} show that CVSS severity and many other CVE properties can be accurately predicted based on only CVE descriptions.
You et al.~\cite{You@2017} extracts vulnerable versions and components from CVE descriptions, and combine them with code commits to generate the proof-of-concept exploits for the CVEs.
Yang et al.~\cite{davidloicse2020veracode} predict the library name fo a CVE based on its description and the Common Platform Enumeration (CPE) dictionary~\cite{CPE}.
Dong et al.~\cite{inconsistance@usenix} extract vulnerable products and versions from CVE descriptions and study their inconsistencies in CVE and NVD.
Different from these works, we study the exploit posts that usually contain much more details about the vulnerabilities and how to exploit them than the concise CVE descriptions.

Our work is the first to summarize the detailed exploit posts into concise CVE descriptions that adhere to the suggested CVE description templates, which opens the door to new vulnerability support and analysis (see Section~\ref{sec:discussion}).
Existing work~\cite{You@2017,inconsistance@usenix} extracts only product, version and component information, while our approach extracts not only these software-specific entities, but also root-cause, attack-vector and impact clauses.
Furthermore, the ExploitDB posts are much more verbose and noisy than CVE descriptions.
Instead of using deep learning based text summarization methods~\cite{liu@emnlp2019,Nallapati@2017,dong@2018acl}, we adopt appropriate NER and QA techniques for different kinds of information based on their data characteristics.
Our experiments confirm the superiority of our approach design and technique choices over existing methods~\cite{inconsistance@usenix,masuyutse2019,liu@emnlp2019}.
In software engineering and security domain, researchers propose software-specific NER techniques~\cite{dehengsaner2016,dehengicsme2016,masuyutse2019,Pandita@usenix2013,Feng@usenix2018,inconsistance@usenix}.
But we are not aware of the use of QA techniques to extract complex clauses in software text.

\section{Conclusion}
This paper conducts the first empirical study to investigate the relationship between the ExploitDB posts and the CVEs.
The study reveals the importance of the ExploitDB posts as an active vulnerability source, and the delay and missing in documenting the published exploits as official CVEs.
Inspired by our study findings, we develop the first approach for composing an informative CVE description from the verbose and noisy exploit posts.
The CVE description is the most important and must-provided information for documenting CVEs.
Our large-scale evaluation confirms the accuracy of the extracted aspects and the quality of the composed CVE descriptions.
In addition to assisting in creating new CVEs, our approach will also be useful for ensuring data consistency within the ExploitDB posts and across heterogeneous vulnerability databases (e.g., ExploitDB and CVE/NVD).

\bibliographystyle{IEEEtran}
\bibliography{IEEEabrv,icse2021citelist}

\end{document}